# Image segmentation by adaptive distance based on EM algorithm

Mohamed Ali Mahjoub 1
Preparatory Institute of Engineer of Monastir
Street Ibn Eljazzar Monastir
Tunisia
Medali.mahjoub@ipeim.rnu.tn

Karim Kalti 2
Faculty of science of Monastir
Street Kairouan - Monastir
Tunisia
Karim.kalti2020@gmail.com

*Abstract*— This paper introduces a Bayesian image segmentation algorithm based on finite mixtures. An EM algorithm is developed to estimate parameters of the Gaussian mixtures. The finite mixture is a flexible and powerful probabilistic modeling tool. It can be used to provide a model-based clustering in the field of pattern recognition. However, the application of finite mixtures to image segmentation presents some difficulties; especially it's sensible to noise. In this paper we propose a variant of this method which aims to resolve this problem. Our approach proceeds by the characterization of pixels by two features: the first one describes the intrinsic properties of the pixel and the second characterizes the neighborhood of pixel. Then the classification is made on the base on adaptive distance which privileges the one or the other features according to the spatial position of the pixel in the image. The obtained results have shown a significant improvement of our approach compared to the standard version of EM algorithm.

*Keywords; EM algorithm, image segmentation, adaptive distance*

## I. INTRODUCTION

Image segmentation is one of the major challenges in image processing and computer vision. When the problem of image segmentation is approached by a process of classification, several methods exist. Recently, finite mixture models have attracted considerable interest for image segmentation. However, the application of finite mixtures model to image segmentation presents some difficulties. For the classical mixture statistical model each pixel must be associated with exactly one class. This assumption may be not realistic. Thus, several methods have been proposed to circumvent this problem. For example, the fuzzy-c mean algorithm has widely been used in image segmentation. Also, some methods mixing fuzzy and statistical model have been developed by Gath. Recently, the Markov Random Field (MRF) models were used with images in an important number of works to add spatial smoothness into the process of image segmentation. This approach provide satisfactory results in many case, but most case the assumption of a single Gaussian distribution typically limits image segmentation accuracy.

The segmentation algorithm developed in this paper is based on a parametric model in which the probability density function of the gray levels in the image is a mixture of Gaussian density functions. This model has received considerable attention in the development of segmentation algorithms and it has been noted that the performance is influenced by the shape of the image histogram and the accuracy of the estimates of the model parameters. However, the model-based segmentation algorithms not allows a good results if the histogram of an image is a poor approximation of a mixture of two Gaussian density functions. The application of this model in image segmentation is, therefore, limited to the images which are a good approximations of Gaussian mixtures with well-defined modes.

Among the techniques of classification, we used in this work models of Gaussian mixtures. An EM algorithm is developed to estimate parameters of the Gaussian mixtures. The finite mixture is a flexible and powerful probabilistic modeling tool. It can be used to provide a model-based clustering in the field of pattern recognition. However, the application of finite mixtures to image segmentation presents some difficulties; especially it's sensible to noise. In this paper we propose a variant of this method which aims to resolve this problem..

The remainder of the paper is organized as follows. Section 2 describes the EM algorithm briefly and the parameter estimation of multi-variate normal densities by the EM algorithm. In Section 3 an unsupervised segmentation algorithm for grey level images is proposed on the basis of the ML estimation. Section 4 presents experimental results to show the performance of the proposed algorithm, and Section 5 concludes by summarizing the paper

## II. SEGMENTATION METHOD BASED ON GMM-EM

The expectation-maximization algorithm (in English Expectation-maximization algorithm), proposed by Dempster, is a class of algorithms that find the maximum likelihood parameters of probabilistic models where the model depends on unobserved variables. The finite mixture of distributions has provided a mathematical approach to the statistical modeling of a wide variety of random phenomena. In the past decades, the extent and the potential of the applications of finite mixture models have widened considerably. In the field of pattern recognition, the finite mixture can be used to provide a model-based clustering. A finite mixture density has the form :

$$f(x|\theta) = \sum_{i=1}^{K} p_i f_i(x|\alpha_i)$$

Where :
- $p_i$ is the proportion of the class $i$ ( $p_i \geq 0$ and $\sum_i p_i = 1$)
- $\alpha_i = (\mu_i, \Sigma_i)$, $\mu_i$ et $\Sigma_i$ are respectively the center and the variance matrix of $k^{th}$ normal component $f(.|\alpha_i)$.

The evaluation of all these parameters can be calculated by maximizing the log-likelihood of the global parameter θ:

$$\ln f(X|\theta) = \sum_{j=1}^{n} \ln f(x_j|\theta) \quad (1)$$

Where $X=(x_1,\ldots,x_n)$, this maximization can be made by the EM algorithm of Dempster. Besides, the k number of components can also be estimated while keeping the value between k=1 and k_sup (k_sup to choose a priori) that minimizes the Bayesian BIC criteria.

$$BIC(K) = -2 \ln f(X|\hat{\theta}_K) + v_K \ln n \quad (2)$$

Where $\hat{\theta}_K$ and $v_K$ are respectively the maximum likelihood estimator and the number of degrees of freedom model.

### A. Classification

Any classification model is defined on the space N of maps from the image domain to the set *N* of classes (each class *n* corresponds to an entity of interest in the scene). Thus each classification $v \in N$ assigns a class $n = v(p) \in N$ to each pixel *p* giving the class of that pixel. By defining a posterior probability distribution on N, and using a suitable loss function, an optimal classification can be chosen. The loss function is more often than not taken to be the negative of a delta function, the resulting estimate then being a maximum a posteriori (MAP) estimate. The posterior distribution is expressed as the (normalized) product of a likelihood, such as the GMM (Gaussian Mixture Models) models that we will discuss in this paper, which gives the distribution of images corresponding to a given class, and a prior probability distribution on the classifications.

### B. Parameters estimation

The EM (expectation-maximization) algorithm is an iterative approach to compute maximum-likelihood estimates when the observations are incomplete. In the mixture density estimation, the information that indicates the component from which the observable sample originates is unobservable. Expectation Maximization (EM) is one of the most common algorithms used for density estimation of data points in an unsupervised setting. In EM, alternating steps of Expectation (E) and Maximization (M) are performed iteratively till the results converge. The E step computes an expectation of the likelihood by including the latent variables as if they were observed, and a maximization (M) step, which computes the maximum likelihood estimates of the parameters by maximizing the expected likelihood found on the last E step. The parameters found on the M step are then used to begin another E step, and the process is repeated until convergence.

The EM algorithm is very used for the research of the parameter achieving the maximum likelihood. The criteria of stop of the algorithm, is either a maximum number of iterations to limit the time of calculation, either a lower mistake. It is put easily in application because it leans on the calculation of the complete data. The EM algorithm requires the initialization of model parameters of Gaussian mixture. The covariance matrix are initialized by the identity matrix, and K mean vectors are initialized by the various centers of Gaussian mixture estimated by the algorithm K-means.

Let's suppose that we have K components in the model of mixtures. The shape of the density of probability of this mixture is given by :

$$f(x|\theta) = \sum_{i=1}^{k} \alpha_i f_i(x|\theta_i) \quad (3)$$

where x is the characteristic vector, $\alpha_i$ is the weight of the mixture as $\sum_{i=1}^{k} \alpha_i = 1$ , $\theta$ represents the parameters $(\alpha_1, \alpha_2, \ldots, \alpha_k, \theta_1, \theta_2, \ldots, \theta_k)$ and $f_i$ the density of the gaussian parameterized by $\theta_i$ that is to say $(\mu_i, \sigma_i)$ :

$$P(x|\theta_i) = \frac{1}{\sqrt{2\pi}\sigma_i} \exp\left\{-\frac{(x-\mu_i)^2}{2\sigma_i^2}\right\} \quad i = 1,2,\ldots,N$$

where $\theta_i = (\mu_i, \sigma_i)$ is the Gaussian mixture distribution parameter.

Assume that the density is derived from a mixture of Gaussians. That is to say that : $f(x) = \sum_{i=1}^{k} p_i \varphi(x, \mu_i, \Sigma_i)$. We will then estimate the parameters by maximizing the likelihood. For this, we must start with a number of K Gaussian fixed a priori and then seek a local maximum in the first order conditions. Thus, the EM algorithm is summarized as follows :

---

Input : H=histogram, k=gaussian number, $\varepsilon$ = Error
Output : model parameters $(p_1, p_2, ..., p_k, \alpha_1, \alpha_2, ...., \alpha_k)$
Streps :
1- Evaluation expectancy (E) :

$$\varphi(i|x_j, \theta) = \frac{p_i f_i(x_j|\alpha_i)}{\sum_{k=1}^{K} p_k f_k(x_j|\alpha_k)}$$

2- Maximization setp (M) : GMM parameters update

$$p_i^{new} = \frac{1}{N} \sum_{j=1}^{N} \varphi(i|x_j, \theta^{old})$$

$$\mu_i^{new} = \frac{\sum_{j=1}^{N} x_j \varphi(i|x_j, \theta^{old})}{\sum_{j=1}^{N} \varphi(i|x_j, \theta^{old})}$$

$$\sum_i^{new} = \frac{\sum_{j=1}^{N} \varphi(i|x_j, \theta^{old})(x_j - \mu_i^{new})(x_j - \mu_i^{new})^T}{\sum_{j=1}^{N} \varphi(i|x_j, \theta^{old})}$$

We stop when $\|\theta^{new} - \theta^{old}\| \leq \varepsilon$

Algorithm 1**:** EM Algorithm

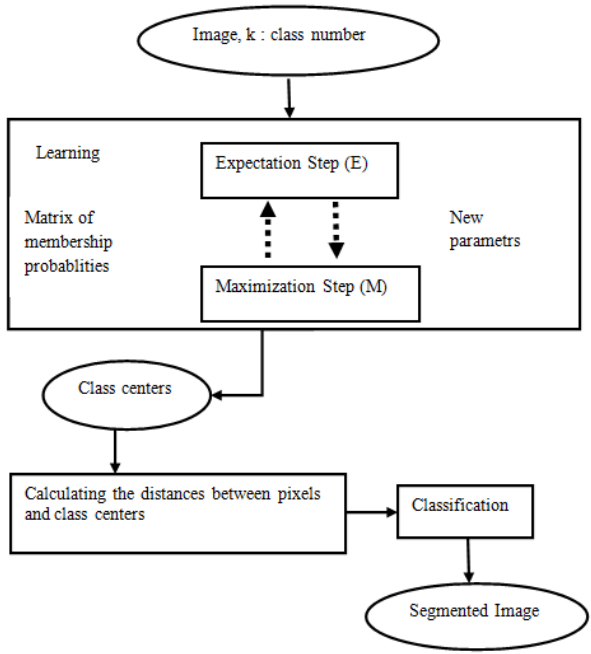

Figure 1: Pixels classification by EM algorithm

However, during the classification phase, we can modify this algorithm to classify a given pixel by calculating the distance to the center of the class instead of a probability calculation. the modified algorithm is called DEM (Distance EM) algorithm (figure 1).

III. LIMITS OF THE GMM SEGMENTATION AND POSSIBLE SOLUTION

The application of standard GMM (using a grey level as a single feature of pixel) yields a good segmentation of pixels inside regions and pixels of contours. However, it includes a bad clustering of noisy pixels of white region. This clustering drawback is essentially due to the only utilization of the intrinsic feature of pixel to be classified (gray level) without take into account the information relative to spatial position in the image. This information turns out to be important in the segmentation context. To overcome this limitation, we propose a solution consisting to include the neighborhood effect of pixel to be classified. To accomplish this effect, we have used in this work the arithmetic average estimator ($\mu$) defined as follows:

$$\mu(xj) = \frac{1}{N} \sum_{k=1}^{N} x_k$$

where $x_j$ designates the pixel to classify, $x_k$ designates a pixel belonging to the analysis window that determines the neighborhood and centered on $x_j$, N designates the number of pixels of the analysis window.

The manner with which will be integrated the describer of neighborhood requires the knowledge of the effect of this last on the process of classification. We present in this paragraph a study of this effect on the noisy and contour pixels. The application of the GMM segmentation algorithm can generate a good classification of the internal pixels to the regions as well as well as the noisy pixels, but produces a deterioration of the contours between the two regions. Indeed, the effect of smoothing introduces by the average generated the attenuation of the noise since its value became enough close to those of its neighbors, thing that drove to its affectation to the same class that these last. This attenuation due to smoothing also affects the pixels of the contours. The gap between the values of these last became less franc and drove toward a bad classification of some among them notably those that have in their setting more of pixels belonging at the neighboring region that to their own region. Of this study we present in figure 2 the advantage and the inconvenient of the use of the grey level and the spatial feature for the noise and the contours classification.

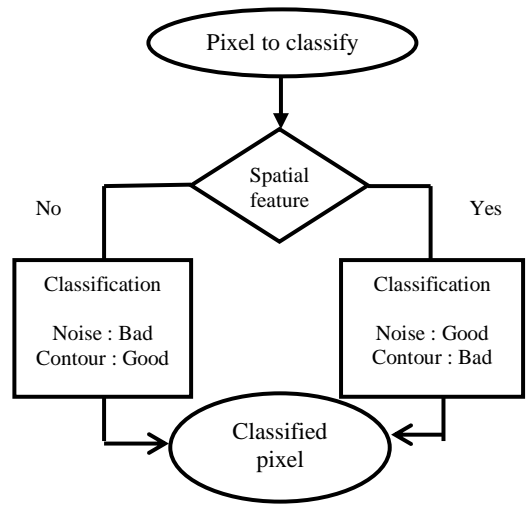

Figure 2 : Spatial feature and pixel classification

## IV. Proposed Approach

The complementarity of the level of gray and the average of the gray levels with regard to the classification can let consider an use joined of these two describers in the image segmentation with the help of the EM method. In this section we present a new version of the EM baptized Adaptive Distance EM (ADEM) that enrolls in this direction. This version tried to take advantage of the advantages of the two aforesaid featrures while avoiding their inconveniences and this while using an adaptive manner one or the other according to the spatial configuration of the pixel to classify.

In the setting of our work, we distinguished four possible spatial configurations for the pixels asking for each a specific choice of the criteria of classification (figure 3). These configurations are the following:.
· PR : Pixel which belongs to a region
· PC : Pixel which belongs to a contour,
· NP : Noisy pixel,
· NNP : Neighbour of Noisy Pixel.

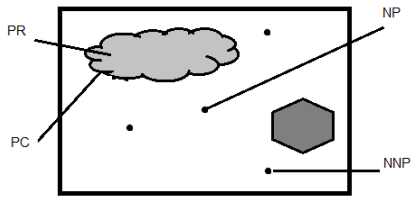

Figure 3 : Types of pixel noise/region/contour

Formally, the different spatial configurations are characterized by the two statistical descriptors presented as follows :
– The standard deviation (σ) characterizes the dynamics of distribution around the pixel to be classified. This attribute is calculated as follows :

$$\sigma(xj) = \sqrt{\frac{1}{N}\sum_{k=1}^{N}(xk - \mu(xj))^2} \quad (4)$$

– The $NCN$ : represents the closest neighbours number in term of grey level regarding the considered pixel. It's defined as follows :

$$NCN(x_j) = card\{x_p \in neighborhood\ (x_j)/\ |x_p - x_j| < S\} \quad (5)$$

S designates a threshold which is generally chosen in year empiric manner. From thesis two descriptors we can characterize the different spatial configurations possible of the pixels. In case of a PR the standard deviation $\sigma$ is generally low, it is null for the constant regions. However the $\sigma$ becoms high for gthe PC, NP and NNP. The distinction between these three configurations is made by using the NCN feature, which is generally low for a NP, moderate for a PC and high for a NNP.

The classification of a PR or a PB must privilege the spatial attribute because the decision must be taken on the basis of the information of its neighborhood. On the other hand the classification of a PC or a NCN must privilege the level of gray of the pixel (NG) respectively to preserve the contours and to avoid the influence of the noise. The choice of the criteria of classification as well as the characterizations of the spatial configurations are summarized in the figure 4.

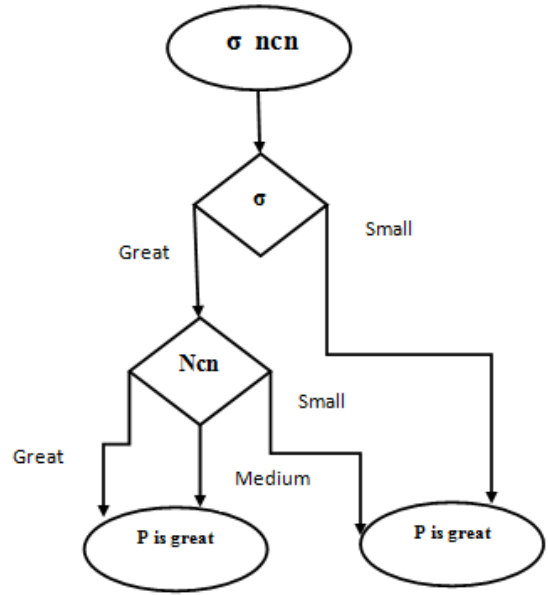

Figure 4 : Influence of $\sigma$ and $NCN$

### A. Proposal of new parameter distance

The DEM algorithm generally uses to measure the similarity between an individual $x_j$ and a class given by its center $v_j$, a distance that grants the same importance to the different attributes took in account in the process of classification. To introduce the adaptive effect as for the selection of the attributes, we propose to use a dynamic distance and weighted derivative of the Euclidian distance. This new distance is given by the equation 6.

$$D(xj, vj) = (1 - pj)(xj^{NG} - vj^{NG})^2 + pj(xj^{Spatial} - vj^{Spatial})^2 \quad (6)$$

D is a bidimensional distance based on the two NG and Spatial attributs. In this equation, the weight pj permits to control the importance of every attribute for the classification of the pixel xj. So if the pj is thigh we privileges the spatial attribute otherwise one privileges the gray level. The term pj must be calculated for every pixel to classify according to its spatial configuration in the image. From the configurations presented in figure 4, we can deduct that the weight pj must be maximized (tender verse 1) when the pixel to classify is a PR or a PB, because the decision of its adherence to the different classes must be taken only on the basis of the spatial attribute. However this pj must be minimized (tender verse 0) in case of a PC or a NNP because the gray level in these cases is going to constitute a good criteria of classification.

### B. Estimation of the spatial weight

The choice of the spatial weight pj is very important for the calculation of the new distance (6). We propose in this paragraph a method of fuzzy evaluation of this weight. For that we use a fuzzy system which processes as entries to linguistic variables of decision $\sigma$ and NCN to give some results on the linguistic variable exit p. While considering the choices of the p in accordance with the spatial configurations, is it possible to define for each of these configurations, a fuzzy rule of the type SO THEN. We can deduct in this case four rules characterizing the relations between the classes of entry ($\sigma$ and NCN) with the corresponding exit class (p) to determine all necessary consequences to calculate the value of p. The linguistic rules that we defined are the following :

R1: *IF $\sigma$ is low THEN p is high*
R2: *IF $\sigma$ is high AND NCN is low THEN p is low*
R3: *IF $\sigma$ is high AND NCN is high THEN p is HIGH*
R4: *IF $\sigma$ is high AND NCN is moderate THEN p is low*

The different membership functions of variables $\sigma$, NCN and P are represented in the figure 5.

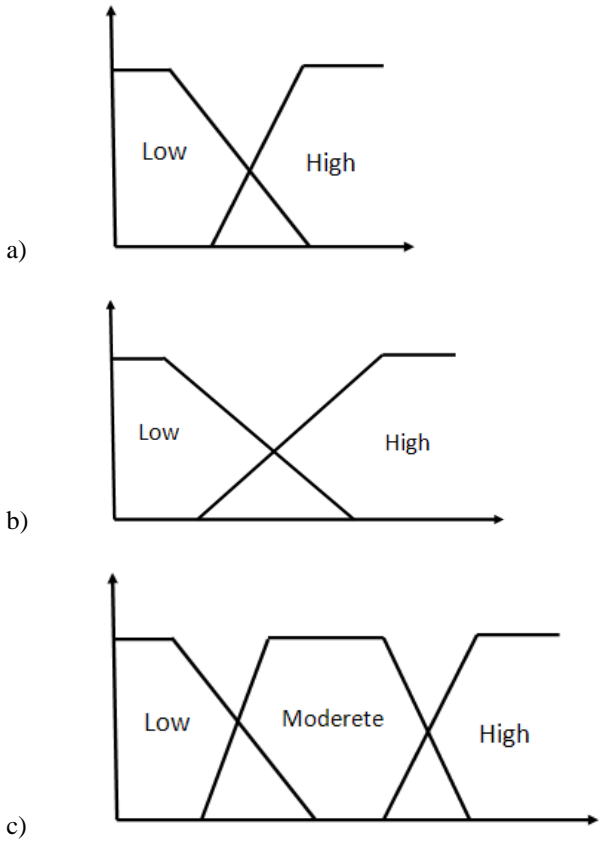

Figure 5 : Membership functions of the linguistic variables used for the estimation of the P

### C. Algorithmic description

Step 1 :
Definition of inputs and outputs.
Input there are two variables $\sigma$ and NCN which characterizes each pixel and the output was the variable P.

Step 2 : Fuzzification or calculating membership degrees of input variables

- Calculating the membership degree of $\sigma$ to the fuzzy set Small (d_sig_S)
- Calculating the membership degree of fuzzy set $\sigma$ to Great (d_sig_G)
- Calculating the membership degree of the fuzzy NCN Small (d_NCN_S)
- Calculating the membership degree of the fuzzy NCN Average (d_NCN_M)
- Calculating the membership degree of the fuzzy NCN Great (d_NCN_G)

Step 3 :
Using the basic rules already declared and using membership degrees computed already, we can calculate the membership degree of the output variable p to the fuzzy set Small (d_p_P) and its degree of membership to all blur Great (d_p_G) and the latter by replacing the logical AND of the basic rules by the MIN function and the logical OR of the MAX function.

d_p_S ← MAX (MIN (d_sig_G,d_NCN_G), MIN (d_sig_G,d_NCN_M))

d_p_G ← MAX ( d_sig_S,, MIN (d_sig_G,d_NCN_S))

The intersection of two lines of equations y1 = d_p_P , y2 = d_p_G and with the curve of the membership function p gives us two surfaces S1 and S2. These two surfaces are none other than a fuzzy value, therefore it must be transformed into real physical quantity.

Step 4 : Defuzzification
There are several methods to calculate the value of the output variable p, we chose the centroïdes method because it is the most accurate:

$$p \leftarrow (S1*X1+S2*X2) / (S1+S2)$$

where : X1 is the center of gravity of the surface S1
X2 is the center of gravity of the surface S2

## V. EXPERIMENT AND RESULTS

In this section, we present the results of the application of the ADEM algorithm. The performance of this algorithm is compared with the standard EM and DEM algorithm. The three techniques are tested on two images, the first is synthetic (panda image) and the second is the MRI cerebral image. The experiences are made in the same conditions (factor of fuzzyfication m = 2 and mistake of convergence = 0.001). Concerning the spatial attributs, the ADEM algorithm uses lthe average calculated on a window of size 3x3 analysis.

The first test applies to the synthetic image named "Panda" (see Figure 6.a). This image contains 3 classes areas clearly identified. The image in Figure 6b shows the result of the application of the standard EM using as criterion classification level of gray. This result shows the limits of this method on classification of noisy pixels. Compared to this, the application of DEM based on the average as attribute classification solves the problem noise as demonstrated in Figure 5.c but leads in cons part segmentation less accurate contours between regions and fine structures (Branches of the tree). The new variant that we proposed is given in figure 5.d. it confirms the good performance compared to DEM and standard EM. Indeed, through the use of the adaptive distance ADEM has secured a compromise that allowed the reduction of noise while producing precise contours.

The second test is about a MRI cerebral image. The segmentation consists in delimiting the three cerebral structures: gray matter (MG), white matter (WM) and the cerebrospinal fluid (CSF). It has particularities bound mainly to the noises and to the effect of the partial volume that one recovers when a pixel having a certain level of gray actually corresponds to a mixture of two or several cloths. This artifact exists mainly at the borders between the cloths MG and WM. The tests are achieved on the image of the corrupt face 6.a by 5% of Gaussian noise.

Table I : Badly classified pixels number contours and inside regions of MRI cerebral image

|         |     | EM   | DEM  | ADEM |
|---------|-----|------|------|------|
| Region  | CSF | 200  | 10   | 4    |
|         | GM  | 500  | 60   | 30   |
|         | WM  | 15   | 5    | 0    |
| Contour | CSF | 500  | 450  | 100  |
|         | GM  | 1120 | 1100 | 190  |
|         | WM  | 450  | 70   | 15   |

The application of the standard EM on MRI image gives some noisy classes and very ridden especially between the two classes MG and MB (figure 7.b). The use of the adaptive distance based EM on the spatial attribute provokes a deterioration of the contours (figure 6.c). while the use of the ADEM (σ=40) permits to reduce the noisy pixel significantly while getting the well identified regions and having continuous and near contours of the reality (figure 6.d). These good performances are confirmed well by the relative statistics to the pixels badly classified. This visual result is supported by the statistics in Table I which gives the number of badly classified pixels within regions and contours.

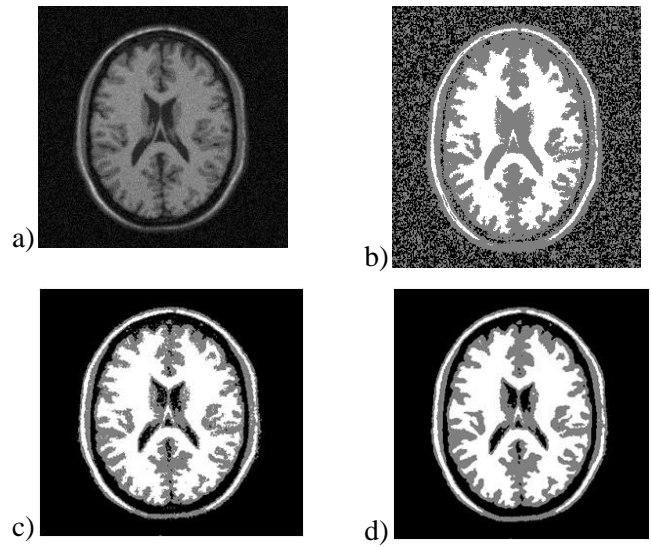

Figure 7 : a) original image b) Result of the standard EM c) Result of the DEM  d) Result of the ADEM (σ=40)

On the other hand, experiments have shown that a value of sigma equal to 40 is quite sufficient for segmenting MRI images (figure 8).

VI.  CONCLUSION

A segmentation algorithm for grey level images was proposed. In the algorithm the image is considered as a mixture of multi-variate normal densities and segmentation is carried out by the ML estimation based on EM algorithm. It's a new approach that we proposed of pixels classification based on a dynamic and weighted similarity distance. The main idea is to use a manner conjoined and adaptive two attributes of classification: level of gray of the pixels and a spatial attribute (the local average). The adaptation must privilege one or the other of these attributes according to the spatial configuration of the pixel to classify. The implementation of this adaptive effect is gotten through the level-headedness of these attributes. The originality of our approach resides in the manner with which is calculated these weights and that takes as a basis on a mechanism of fuzzy inference. The new distance that we proposed thus permitted to get a new variant of the EM method that is adapted more to the segmentation of images. It has been confirmed by the tests that we achieved on a cerebral MRI image. Indeed, these last showed a clean improvement of the performances of our approach notably in relation to the version classic EM with regard to the hardiness in relation to the noise and the precision of the contours gotten.

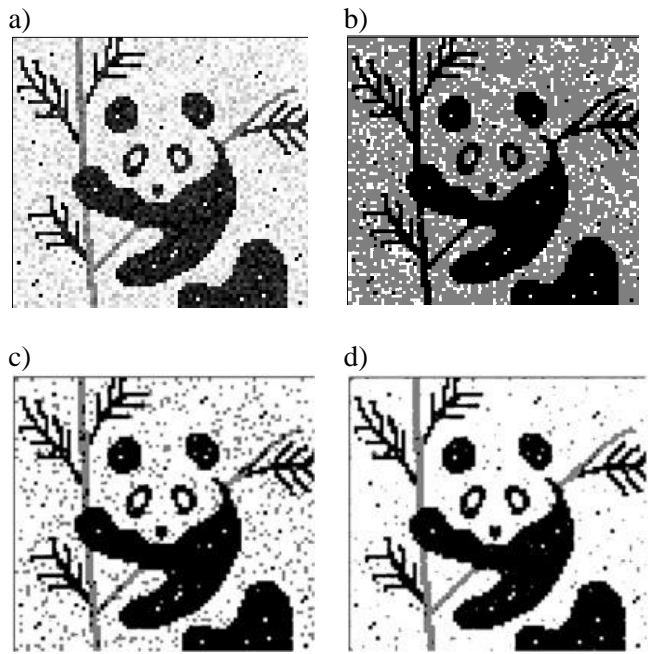

Figure 6 : a) original image b) Result of the standard EM c) Result of the DEM  d) Result of the ADEM (σ=40)

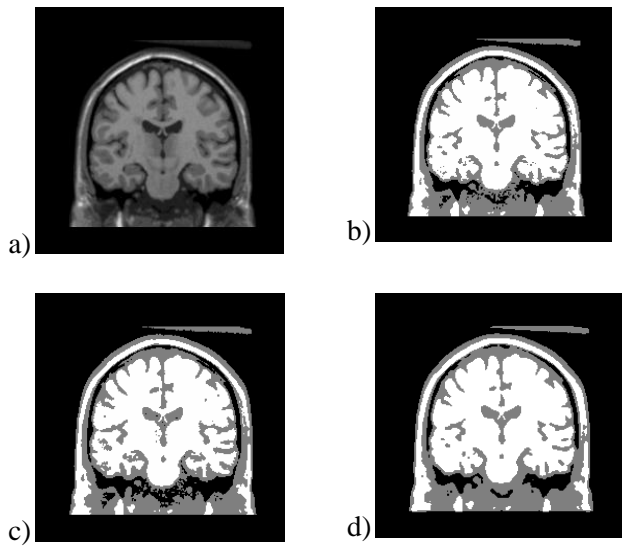

Figure 8 : Figure 7 : a) original image b) Result of the ADEM (σ =15) c) Result of the ADEM (σ=40) d) Result of the ADEM (σ=70)

AUTHORS PROFILE

**Dr Mohamed Ali Mahjoub** is an assistant professor in the department of computer science at the Preparatory Institute of Engineering of Monastir. His research interests concern the areas of Bayesian Network, Computer Vision, Pattern Recognition, Medical imaging, HMM, and Data Retrieval. His main results have been published in international journals and conferences.

**Dr Karim kalti** is an assistant professor in the department of computer science at the faculty of science of Monastir (Tunisia). His research interests concern the areas Computer Vision, Pattern Recognition, Medical imaging and Data Retrieval. His main results have been published in international journals and conferences.